\documentclass[10pt,twocolumn,letterpaper]{article}

\usepackage{cvpr}
\usepackage{times}
\usepackage{epsfig}
\usepackage{graphicx}
\usepackage{amsmath}
\usepackage{amssymb}
\usepackage{floatrow}
\usepackage{caption}
\usepackage{url}

\usepackage{booktabs}
\usepackage[numbers,sort&compress]{natbib}
\usepackage{color}
\usepackage[normalem]{ulem}

\newfloatcommand{capbtabbox}{table}[][\FBwidth]

\cvprfinalcopy 

\def\cvprPaperID{1959} 

\ifcvprfinal\fi
\begin{document}

\title{Learning Compositional Visual Concepts with Mutual Consistency}

\author{Yunye Gong$^{1}$, Srikrishna Karanam$^{3}$, Ziyan Wu$^{3}$, Kuan-Chuan Peng$^{3}$, Jan Ernst$^{3}$, and Peter C. Doerschuk$^{1,2}$\\
$^{1}$School of Electrical and Computer Engineering, Cornell University, Ithaca NY\\
$^{2}$Nancy E. and Peter C. Meinig School of Biomedical Engineering, Cornell University, Ithaca NY\\
$^{3}$Siemens Corporate Technology, Princeton NJ\\
{\tt\small \{yg326,pd83\}@cornell.edu,\{first.last\}@siemens.com}
}


\twocolumn[{%
\renewcommand\twocolumn[1][]{#1}%
\maketitle
\vspace{-4em}
\begin{center}
    \centering
    \includegraphics[scale=0.75]{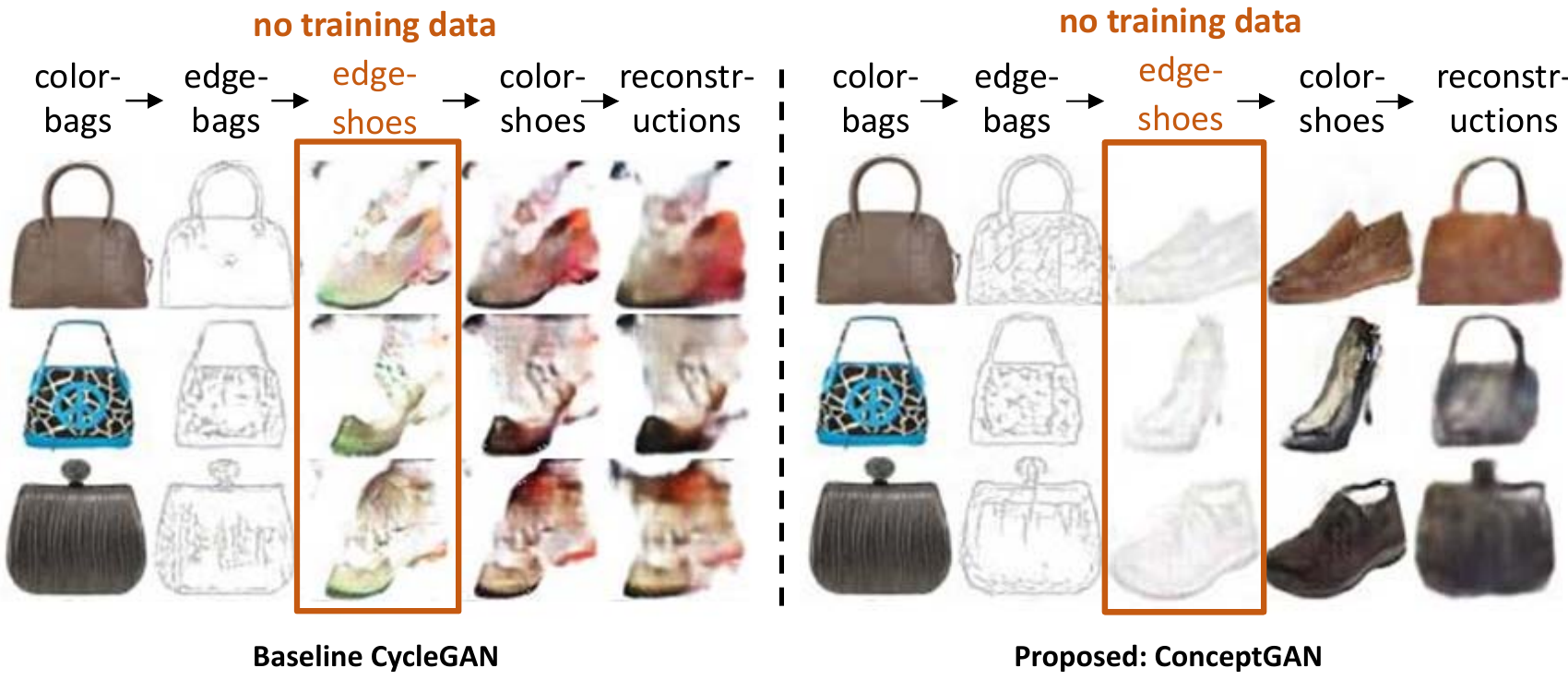}
    \captionof{figure}{We propose ConceptGAN, a framework that can jointly learn, transfer and compose concepts to generate semantically meaningful images, even in subdomains with no training data (highlighted) while the state-of-the-art methods such as CycleGAN~\cite{CycleGAN} fail to do so.}
    \label{fig:teaser}
\end{center}%
}]

\begin{abstract}
\vspace{-1em}
Compositionality of semantic concepts in image synthesis and analysis is appealing as it can help in decomposing known and generatively recomposing unknown data. For instance, we may learn concepts of changing illumination, geometry or albedo of a scene, and try to recombine them to generate physically meaningful, but unseen data for training and testing. In practice however we often do not have samples from the joint concept space available: We may have data on illumination change in one data set and on geometric change in another one without complete overlap. We pose the following question: How can we learn two or more concepts jointly from different data sets with mutual consistency where we do not have samples from the full joint space? We present a novel answer in this paper based on cyclic consistency over multiple concepts, represented individually by generative adversarial networks (GANs). Our method, ConceptGAN, can be understood as a drop in for data augmentation to improve resilience for real world applications. Qualitative and quantitative evaluations demonstrate its efficacy in generating semantically meaningful images, as well as one shot face verification as an example application.

\end{abstract}

\section{Introduction}
\thispagestyle{empty}
In applications such as object detection and face recognition, a large set of training data with accurate annotation is critical for the success of modern deep learning-based methods. However, collecting and annotating such data can be a laborious or even an essentially impossible task. Conventional data augmentation techniques typically involve either manual effort or simple transformations such as translation and rotation of the available data, and may not result in semantically meaningful data samples.

\begin{figure}[!t]
\begin{center}
\includegraphics[scale=0.55]{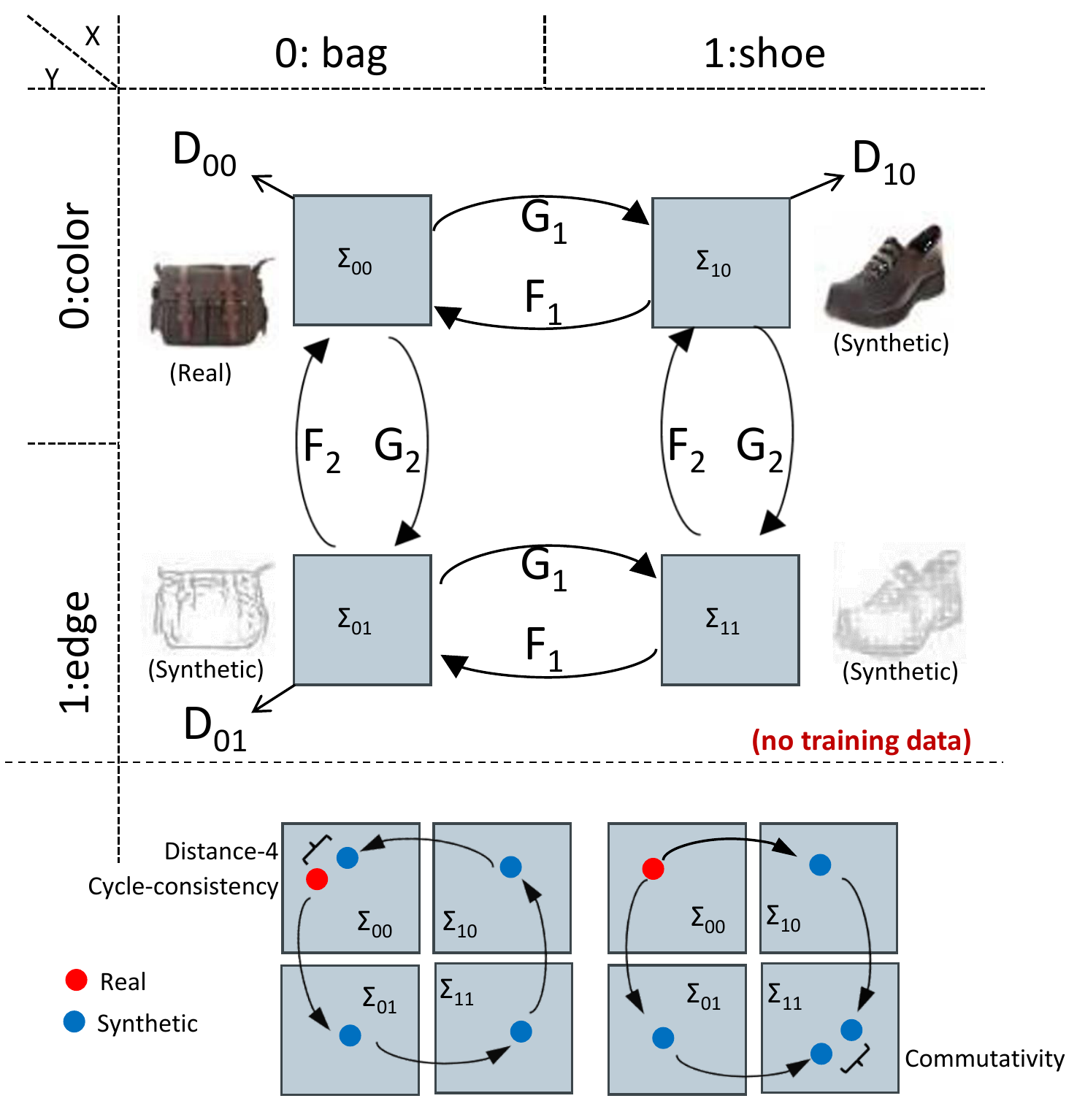}
\end{center}
   \caption{The proposed concept learning approach: Four-vertex cyclic graph for joint learning of two latent concepts.}
\vspace{-1.5em}
\label{fig:cyclicmodel}
\end{figure}

Recently, generative models have been shown to successfully synthesize unseen data samples, such as image-to-image translation and CycleGAN \cite{CycleGAN,DiscoGAN}. Given sufficient training data, these allow us, for instance, to translate from an image of a textured handbag to a corresponding visually convincing image of a shoe with the same texture, or from a color image of a handbag to a consistent line drawing of a handbag. Starting with this limitation of learning one concept at a time, naturally one would like to continue learning more concepts to generate a wider variety of data. However, samples from the joint distribution, in our simple case of line drawings of shoes, may not be available for training. Going beyond two concepts, the joint concept space certainly becomes exponential and unfeasible for gathering data. As shown in Figure~\ref{fig:teaser}, it is difficult to directly compose separately trained CycleGAN mappings in a semantically meaningful way to synthesize plausible images in the subdomains with no training data. For example shape-varying mappings trained with color images may fail to translate images in the line drawing domain. As an answer to this challenge, we make compositionality a principled and explicit part of the model while learning individual concepts. We achieve this by regularizing the learning of the individual concepts by enforcing consistency of concept composition. In our earlier example, this implies enforcing cyclic consistency of applying bag to shoe, color to line drawing, and their corresponding inverses, resulting in a cycle of four concept shifts (Figure~\ref{fig:cyclicmodel}). In general, we enforce consistency over multiple closed paths in the underlying graph. The benefits of this are twofold: (a) we ensure that the concepts are mutually consistent in the sense of not impacting their mutual forward and inverse generation capability, and (b) we can optimize the resulting cost function irrespective of whether data samples from the joint concept space are available. In fact, we focus on the case where no data is available from one joint concept space (e.g., line drawings of shoes) and demonstrate that we can nevertheless generate meaningful samples from it. This paper focuses primarily on the simplest case of our framework with two-concept cycles. 

While not strictly necessary, we assume that the application of concepts is commutative, yielding a set of symmetric cycle consistency constraints. As it is notoriously difficult to gauge the performance of novel image synthesis, we use a surrogate task, face verification, for performance evaluation and demonstrate how a black-box baseline system can be improved by data augmentation. In summary:

\begin{itemize}
\setlength\itemsep{-0.3em}
\item We propose a principled framework for learning pairwise visual concepts from partial data with mutual consistency. 
\item We demonstrate that via joint learning, transfer and composition of concepts, semantically meaningful image synthesis can be achieved over a joint latent space with incomplete data, for instance from a subdomain where no data is available at training time.
\item We demonstrate a scalable framework for efficient data augmentation where multiple concepts learned in a pair-wise fashion can be directly composed in image synthesis. 
\item Using face verification as a surrogate problem, we show how the proposed method can be used as a framework to perform conditional image synthesis, helping improve face verification accuracy.
\item We provide a scheme for building iterative solutions for an arbitrary number of concepts as a generalization.
\end{itemize}

\section{Related work}

The challenge of data scarcity has been addressed in various computer vision research~\cite{data_scar1, data_scar2, kunpeng}. In particular, data augmentation techniques have been utilized to improve the training performance especially for deep learning-based methods~\cite{flipping, flipping2, cropping, pose_est}. Conventional approaches mostly rely on simple transformations such as rotation~\cite{bestprac}, random cropping~\cite{cropping}, random flipping~\cite{flipping,flipping2,cropping} and altering RGB channel intensities~\cite{altRGB}.~The amount of new information introduced in such operations is limited as no latent manipulation (e.g., varying the illumination) is involved.

Generative adversarial networks (GAN)~\cite{GAN} provide an efficient tool to augment data with virtual samples~\cite{RenderGAN, deraining, zizhao}. In GAN, plausible yet unseen images are generated by matching the synthetic sample distribution to the real data distribution. The adversarial idea has been successfully applied to the transformation across image domains. Isola et al.~\cite{pix2pix} propose the pix2pix framework, which adapts a conditional GAN~\cite{cGAN} to map images from the input to output domain given paired training data. Various strategies have been utilized to tackle the problem with unsupervised data, such as using weight-sharing between adversarial networks to learn the joint distribution across domains~\cite{CoGAN, unsup_im2im} and using an additional regularization loss term which minimizes a similarity distance between the inputs and the outputs~\cite{unsup_crossdomain, unsup_pixel, simGAN}. 

In particular, Zhu et al.~\cite{CycleGAN} propose CycleGAN, which extends the pix2pix~\cite{pix2pix} framework by introducing additional cycle-consistency constraints to simultaneously learn a pair of forward and backward mappings between two domains given unpaired training data. Similar unsupervised learning ideas are also proposed in the DiscoGAN~\cite{DiscoGAN} and the DualGAN~\cite{DualGAN}. Following the cycle-consistency formulation, Liang et al.~\cite{ContrastGAN} focus on editing high-level semantic content of objects while preserving background characteristics. In these prior works, however, translation mappings learned in each experiment depend on specific training distributions, and therefore can not be easily transferred or semantically composed without extra training experiments. 

Another group of generative model-based approaches seek to learn the disentangled latent representations~\cite{VAE, semisup, AAE, inverse, attribute2image} where the semantic perturbation can then be expressed via the vector arithmetic~\cite{word2vec, image2vec}. Various recent efforts have successfully combined the representation learning with adversarial networks in applications such as conditional image synthesis~\cite{auxiliary, rev2_1, rev2_2, rev2_3, rev2_4}. Chen et al.~\cite{InfoGAN} adopt an unsupervised approach by maximizing the mutual information between code space input and output observations. Fu et al.~\cite{CDD} perform conditional image synthesis given training data only supervised in one (source) domain via joint feature disentanglement and adaption. Lu et al.~\cite{cond_cyclegan} and Kim et al.~\cite{attribute_transfer} both propose models on top of a cycle-consistency formulation~\cite{CycleGAN}. 

While these works can provide plausible image synthesis conditional on attribute manipulations, the discussions are still under the assumption that training data are available over the joint latent space and have no accommodation for the challenge of the data scarcity.
Unlike prior works, the proposed ConceptGAN captures image space mappings that correspond to commutative shifts in the underlying latent space. In each experiment, we jointly learn, transfer and compose such concepts to synthesize images over joint latent space including a subdomain missing at the training stage. Given the transferability of such learned concepts, our technique also paves a way for a principled framework to generalize to multiple concepts where new concepts can be learned incrementally without looking at past data.

\section{Model formulation}
\label{sec:model}

We propose ConceptGAN, a concept learning framework aimed at recovering the joint space information given missing training data in one subdomain. As illustrated in Figure~\ref{fig:cyclicmodel}, the basic unit of the framework is modeled as a four-vertex cyclic graph, where a pair of latent concepts is jointly learned. Each vertex refers to a subdomain $\Sigma_{XY}$ with binary latent labels $X$ and $Y$ and corresponding training samples $\{\sigma_{XY}^i\}_{i=1}^{N_{XY}}\in\Sigma_{XY}$, where $N_{XY}$ denotes the number of training samples in the subdomain $\Sigma_{XY}$. The variation over each latent concept is learned as a pair of forward and inverse mappings, $(G_i,F_i)_{i=1,2}$ between subdomains. For example, $G_1 : \Sigma_{X=0,Y}\rightarrow \Sigma_{X=1,Y}$ and $F_1 : \Sigma_{X=1,Y}\rightarrow \Sigma_{X=0,Y}$ define the variation over concept $X$. In particular, no pairwise correspondence is required for data samples between any two subdomains and our goal is to generate realistic synthetic samples over all four subdomains under the assumption that no training samples are available in one of the subdomains. In the following discussion, we assume that the subdomain $\Sigma_{11}$ has no training data (i.e. $N_{11}=0$). An adversarial discriminator $D_{XY}$ is introduced at each of the three subdomains $\Sigma_{00}$, $\Sigma_{01}$ and $\Sigma_{10}$ to tell synthetic data and real data apart. We further extend cycle-consistency constraints used in the CycleGAN~\cite{CycleGAN} and introduce a commutative loss to encourage learning transferable and composable concept mappings.

\subsection{Adversarial loss}

The adversarial loss~\cite{GAN} is applied to each of the three subdomains where real data is available during training, which encourages learning mappings between adjacent subdomains to generate realistic samples. Let $P_{XY}$ denote the underlying distribution of the real data in subdomain $\Sigma_{XY}$. For generator $G_1$ and discriminator $D_{10}$, for example, the adversarial loss is expressed as:
\begin{equation}
\begin{split}
\mathcal{L}_{adv}(G_1,D_{10},\Sigma_{00},\Sigma_{10}) {}=
\mathbb{E}_{\sigma_{10}\sim P_{10}}[\log D_{10}(\sigma_{10})]\\
+\mathbb{E}_{\sigma_{00}\sim P_{00}}[\log(1- D_{10}(G_1(\sigma_{00})))]
\end{split}
\end{equation}

where the generator $G_1$ and discriminator $D_{10}$ are learned to optimize a minimax objective such that
\begin{equation}
G_1^*=\arg\min_{G_1}\max_{D_{10}}\mathcal{L}_{adv}(G_1,D_{10},\Sigma_{00},\Sigma_{10})
\end{equation}
Similarly we define $\mathcal{L}_{adv}(G_2,D_{01},\Sigma_{00},\Sigma_{01})$, $\mathcal{L}_{adv}(F_1,D_{00},\Sigma_{10},\Sigma_{00})$, and $\mathcal{L}_{adv}(F_2,D_{00},\Sigma_{01},\Sigma_{00})$ for $G_2$, $F_1$ and $F_2$ respectively. The overall adversarial loss $\mathcal{L}_{ADV}$ is the sum of these four terms.


\subsection{Extended cycle-consistency loss}
In Zhu et al.~\cite{CycleGAN} a pairwise cycle-consistency loss is proposed to encourage generators to learn bijectional mappings between two distributions. Let $\mathcal{L}_{CYC2}$ denote the sum of all such pairwise (i.e., distance-2) cycle consistency losses adopted in the cyclic model, where six terms are included: (1) both forward cycle-consistency and backward cycle-consistency~\cite{CycleGAN} between pairs $(\Sigma_{00},\Sigma_{01})$ and $(\Sigma_{00},\Sigma_{10})$ and (2) only forward cycle-consistency between pairs  $(\Sigma_{01},\Sigma_{11})$ and $(\Sigma_{10},\Sigma_{11})$. Such consistency constraints can naturally be extended to potentially any closed walks in the cyclic graph and thus further reduce the space of possible mappings. In particular, the difference between training data samples and image samples reconstructed via walking through all four vertices from either direction is minimized. For example, for any data sample $\sigma_{00}$ in subdomain $\Sigma_{00}$, a distance-4 cycle consistency constraint is defined in the clockwise direction $(F_2 \circ F_1 \circ G_2 \circ G_1)(\sigma_{00})\approx \sigma_{00}$ and in the counterclockwise direction $(F_1\circ F_2\circ G_1\circ G_2)(\sigma_{00})\approx \sigma_{00}$. Such constraints are implemented by the penalty function:
\begin{align}
\mathcal{L}_{cyc4}&(G,F,\Sigma_{00}) \nonumber \\
&= \mathbb{E}_{\sigma_{00}\sim P_{00}}[\Vert (F_2\circ F_1\circ G_2\circ G_1)(\sigma_{00})-\sigma_{00}\Vert_1] \nonumber \\
&+\mathbb{E}_{\sigma_{00}\sim P_{00}}[\Vert (F_1\circ F_2\circ G_1\circ G_2)(\sigma_{00})-\sigma_{00}\Vert_1].
\end{align}
Similarly, we define $\mathcal{L}_{cyc4}(G,F,\Sigma_{01})$ and $\mathcal{L}_{cyc4}(G,F,\Sigma_{10})$ considering the case where the original image is in subdomain $\Sigma_{01}$ and $\Sigma_{10}$ respectively. Let $\mathcal{L}_{CYC4}$ denotes the sum of these three terms. The overall cycle consistency loss $\mathcal{L}_{CYC}=\mathcal{L}_{CYC2}+\mathcal{L}_{CYC4}$.



\subsection{Commutative loss}
Adversarial training in Zhu et al.~\cite{CycleGAN} learns mappings that capture sample distributions of training data and therefore are not easily transferable to input data that follows a different distribution without a second training, which may lead to weak compositionality. In order to encourage the model to capture semantic shifts, which correspond to commutative operators such as addition and subtraction in latent space, we enforce a commutative property for concept composition such that starting from one data sample, similar outputs are expected after applying concepts in different orders. For example, for any data sample $\sigma_{00}$ in subdomain $\Sigma_{00}$, we introduce a constraint $(G_2\circ G_1)(\sigma_{00})\approx (G_1\circ G_2)(\sigma_{00})$ implemented by the penalty function:
\begin{align}
\mathcal{L}_{comm}&(G_1,G_2,\Sigma_{00}) \nonumber \\
=&\mathbb{E}_{\sigma_{00}\sim P_{00}}[\Vert (G_2\circ G_1)(\sigma_{00})-(G_1\circ G_2)(\sigma_{00})\Vert_1]
\end{align} $\mathcal{L}_{comm}(G_1,F_2,\Sigma_{01})$ and $\mathcal{L}_{comm}(F_1,G_2,\Sigma_{10})$ are defined in a similar way by considering original image in subdomains $\Sigma_{01}$ and $\Sigma_{10}$. The overall commutative loss $\mathcal{L}_{COMM}$ is the sum of the three terms.


\begin{figure*}[!htb]
\begin{center}
\includegraphics[scale=0.5]{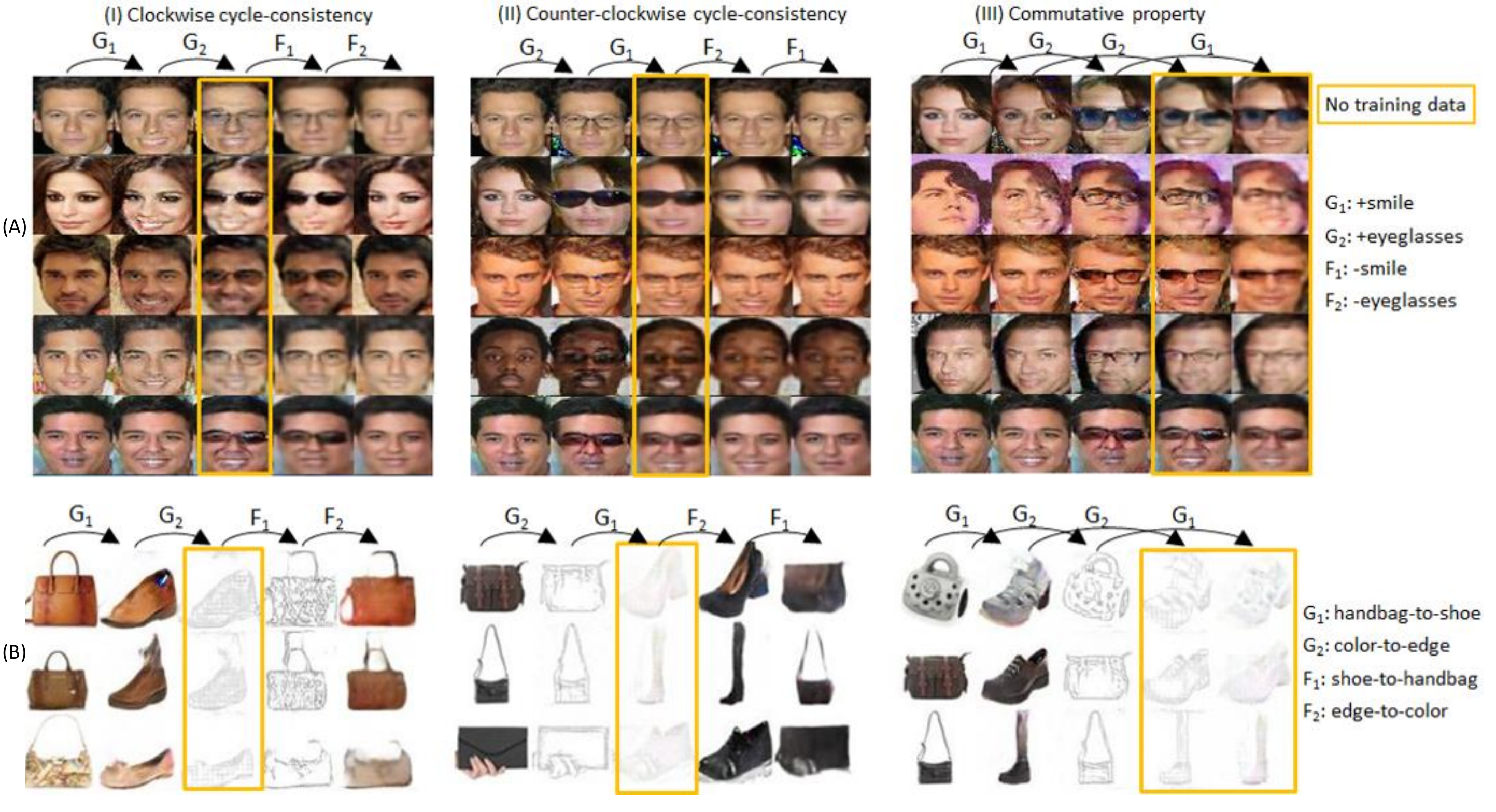}
\end{center}
   \caption{Image translation and synthesis conditional on concepts:  (A) ``smile" and "eyeglasses"; (B) ``handbag vs. shoe" and ``color vs. edge". Each panel in column (I) demonstrates the clockwise cycle consistency where $\sigma_{00}$, $G_1(\sigma_{00})$, $(G_2\circ G_1)(\sigma_{00})$, $(F_1\circ G_2\circ G_1)(\sigma_{00})$, $(F_2\circ F_1\circ G_2\circ G_1)(\sigma_{00})$ are shown in sequence, from left to right. Each panel in column (II) demonstrates the counter-clockwise cycle consistency  where $\sigma_{00}$, $G_2(\sigma_{00})$, $(G_1\circ G_2)(\sigma_{00})$, $(F_2\circ G_1\circ G_2)(\sigma_{00})$, $(F_1\circ F_2\circ G_1\circ G_2)(\sigma_{00})$ are shown in sequence, from left to right. Each panel in column (III) demonstrates the commutative property of the concept composition where $\sigma_{00}$, $G_1(\sigma_{00})$, $G_2(\sigma_{00})$, $(G_2\circ G_1)(\sigma_{00})$, $(G_1\circ G_2)(\sigma_{00})$ are shown in sequence, from left to right. Synthesis results obtained in the subdomains where no training data is available are highlighted in yellow boxes.}
\vspace{-1.5em}
\label{fig:3panels}
\end{figure*}


\begin{figure*}[!htb]
\begin{center}
\includegraphics[scale=0.75]{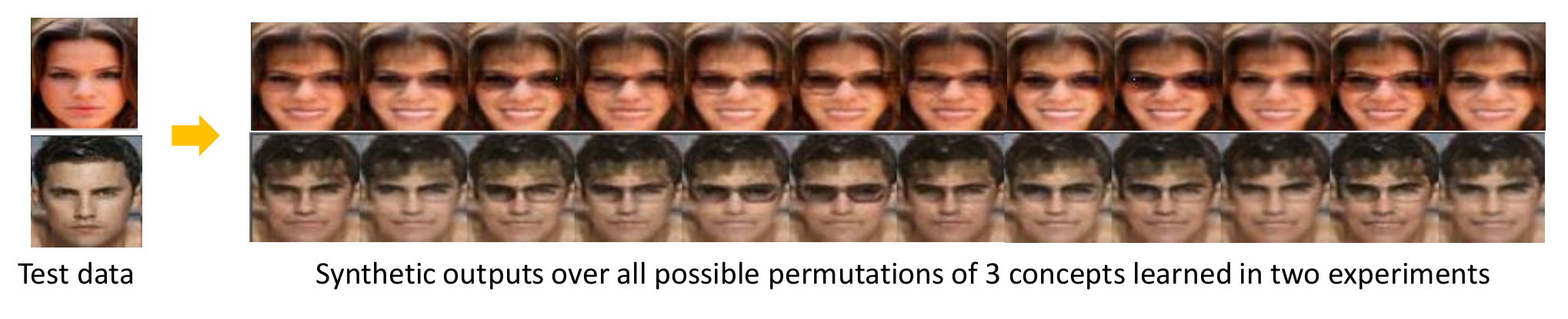}
\end{center}
   \caption{Image synthesis in a zero-shot subdomain by composing three concepts (smile, eyeglasses, bangs) learned in two separate experiments. Concept mappings with respect to "eyeglasses" is learned in each of two experiments therefore $2\times(3!)=12$ different compositions of mappings available to translate images labeled as (no smile, no eyeglasses, no bangs) to the target subdomain.}
\vspace{-2em}
\label{fig:3concept}
\end{figure*}


\begin{figure*}[!htb]
\begin{center}
\includegraphics[scale=0.5]{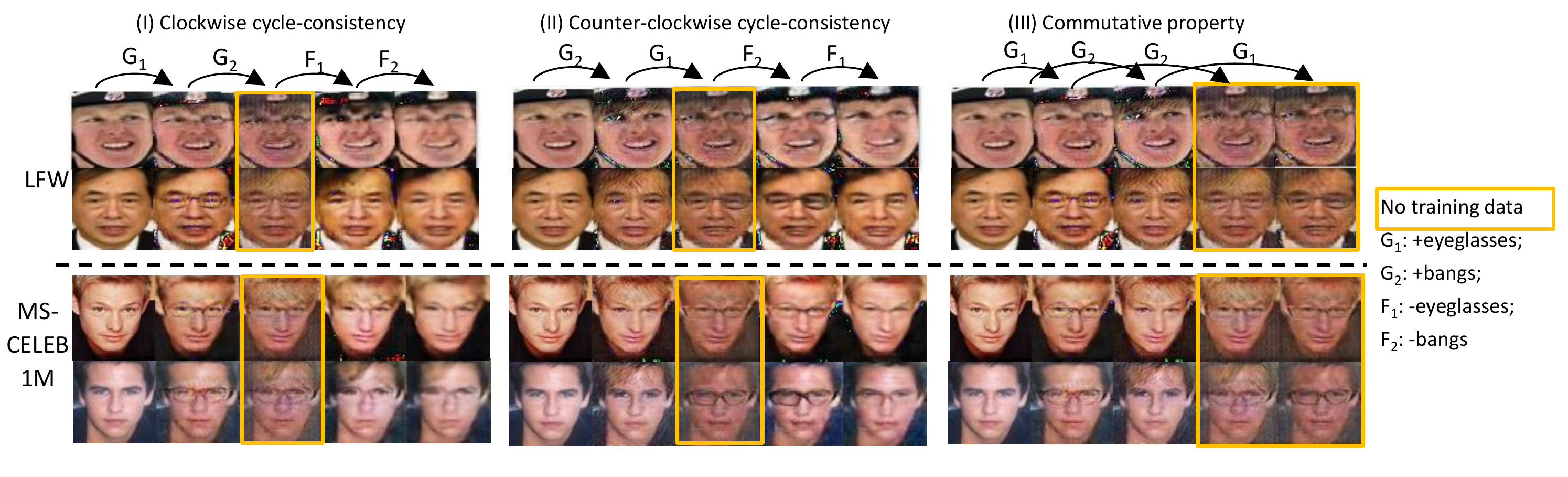}
\end{center}
   \caption{Transfer of learned concepts: Image translation and conditional synthesis on face attributes ``eyeglasses" and ``bangs" via direct application of models trained by CelebA data~\cite{celebA} on independent test datasets MS-Celeb-1M~\cite{msceleb} (top) and LFW~\cite{lfw} (bottom).}
\vspace{-1em}
\label{fig:msceleb-lfw}
\end{figure*}



\subsection{Overall loss function}
\label{sec:overallLoss}
The overall loss function is expressed as:\begin{align}
\mathcal{L}(G,F,D,\Sigma)=\mathcal{L}_{ADV}+\lambda\mathcal{L}_{CYC}+\mu\mathcal{L}_{COMM}
\end{align} with weight parameters $\lambda$ and $\mu$. The generators are learned as the solutions of a minimax problem:\begin{align}
G^*,F^*=\arg\min_{G,F}\max_{D}\mathcal{L}(G,F,D,\Sigma).
\end{align}

\subsection{Composition of multiple concepts}
\label{sec:multiple}
In each experiment, two concepts are jointly trained via the proposed cyclic model shown in Figure~\ref{fig:cyclicmodel}, where synthetic images are generated in all four sudomains. In particular, by composing the pair of concepts, plausible images are synthesized in subdomain $\Sigma_{11}$ where we assume no training data is available. Such image synthesis mechanism can be generalized by considering the composition of multiple concepts. For example, we demonstrate in Figure~\ref{fig:3concept}, that by directly combining two pairs of concepts learned in separate experiments, plausible images can be generated over three dimensional latent space, including a subdomain where no training data is available in either of the experiments, which suggests that the proposed system can be scaled up with linearly increased complexity via direct composition of concepts learned in pairwise fashion.

\subsection{Implementation details} \label{implementation}
For all discriminators, we use the architecture similar to Kim et al.~\cite{DiscoGAN} which contains 5 convolution layers with $4 \times 4$ filters. Compared to the PatchGAN used in Zhu et al.~\cite{CycleGAN}, the discriminator network takes 64x64 input images and output a scalar from the sigmoid function for each image. For all the generators, we use the architecture adapted from Zhu et al~\cite{CycleGAN}, which contains 2 convolution layers with stride 2, 6 residual blocks and 2 fractionally-strided convolution layers with stride $\frac{1}{2}$. We use Adam optimizer \cite{adam} with an initial learning rate of 0.0002 at the first 150 epochs, followed by a linearly decaying learning rate for the next 150 epochs as the rate goes to zero. For experiments in Section~\ref{experiments}, we set $\mu=\lambda=10$ and we also include an identity loss component~\cite{CycleGAN} with weight 10.


\section{Experiments}


\subsection{Conditional image synthesis} \label{experiments}
Image synthesis experiments are performed each corresponding to the manipulation over two concepts. In Figure~\ref{fig:3panels}, column (I), (II) and (III) demonstrate the clockwise cycle-consistency, the counter-clockwise cycle-consistency and the commutative property of the concept composition respectively. Given real testing images shown at the leftmost in each panel, plausible synthetic data are generated with correct semantic variation in each subdomain, including the subdomain where no training data is available.
\\\textbf{Concept learning with face images}  ~Figures~\ref{fig:3panels} (A) and Figure~\ref{fig:3concept} show the results of applying proposed method on face images. The concept learning models are trained and tested on CelebA dataset~\cite{celebA}. In the experiment concerning the concepts ``smile" and ``eyeglasses" (Figure~\ref{fig:3panels}(A)), 4851,3945 and 4618 images with attribute labels (no smile, no eyeglasses), (no smile, with eyeglasses) and (with smile, no eyeglasses) are used at the training stage for subdomains $\Sigma_{00}$, $\Sigma_{01}$ and $\Sigma_{10}$ respectively. 
Figure~\ref{fig:3concept} presents the results of directly composing three concepts learned in two separate experiments described in Section~\ref{sec:multiple}. Synthetic images are generated in the subdomain with labels (with smile, with eyeglasses, with bangs) where no training data is available in either experiment. It is shown that the proposed method can thus be generalized to manipulation over higher dimensional latent spaces.
\\\textbf{Transfer of learned concepts} Here, we qualitatively demonstrate the transferability of the concepts learned by ConceptGAN on different datasets not used at all during training. Figure~\ref{fig:msceleb-lfw} presents the results of this experiment of direct transfer of the learned concept pair to independent test sets. Concepts "eyeglasses" and "bangs" are trained with CelebA~\cite{celebA} dataset and tested on datasets LFW~\cite{lfw} and MS-CELEB-1M~\cite{msceleb} respectively.
\\\textbf{Concept learning of shape and texture}
~Figure~\ref{fig:3panels} (B) shows the results of applying the proposed method on images concerning the concepts ``handbag vs. shoe" (shape variation) and ``photo vs. edge" (texture variation). Without taking advantage of the paired labels, we use the ``edges2shoes" and ``edges2handbags" dataset from ``pix2pix"~\cite{pix2pix} dataset. 5124, 5124 and 4982 images with attribute labels (color, handbag), (edge, handbag) and (color, shoe) are used at the training stage for subdomains $\Sigma_{00}$, $\Sigma_{01}$ and $\Sigma_{10}$ respectively. Given no training data, synthetic line drawing (``edge") images are generated for shoes. 
The importance of simultaneously learning and transferring concept mappings is demonstrated in comparison to results of direct composition of separately trained CycleGAN units~\cite{CycleGAN} in Figure~\ref{fig:teaser}. In particular, the mappings trained via baseline CycleGAN with images in subdomains $\Sigma_{00}$ and $\Sigma_{10}$ are restricted to training distributions and therefore fail to preserve the correct texture information when directly transferred to input images in subdomain $\Sigma_{01}$ \footnote{Additional results of the experiments, including with other concepts, can be found in the supplementary material.}.%



\section{Quantitative evaluations}
We provide quantitative performance evaluations of our proposed concept learning framework for two different tasks: attribute classification and face verification.

\subsection{Attribute classification}

In this section, our goal is to quantitatively demonstrate the importance of simultaneously learning and transferring concept mappings as opposed to learning and composing concepts separately via a single CycleGAN unit. To this end, we perform, and report results of, several classification experiments. Specifically, we employ the following evaluation protocol: (a) We use data in subdomains $\Sigma_{00}$, $\Sigma_{01}$ and $\Sigma_{10}$ to learn concept mappings and automatically synthesize data in the subdomain $\Sigma_{11}$ using the proposed concept learning model. We then use the generated images in the subdomain $\Sigma_{11}$ and perform a two-class classification experiment on each of the concepts. (b) We repeat the experiment described above, but now data in $\Sigma_{11}$ is generated as composition of two independently learned CycleGAN units, i.e., we learn one CycleGAN for the $\Sigma_{00}\implies \Sigma_{10} $ mapping and another CycleGAN for the $\Sigma_{00}\implies \Sigma_{01} $ mapping. Given data in $\Sigma_{00}$, we then compose the two learned mappings to synthesize data in $\Sigma_{11}$. We use the same network architecture to train the separate CycleGAN unit as described in Section~\ref{implementation}.

\begin{table}
\centering
\setlength{\tabcolsep}{3pt}
\scalebox{0.92}{
\begin{tabular}{l@{\hspace{0.2cm}}c@{\hspace{0.2cm}}c@{\hspace{0.2cm}}c}
\toprule
Classifier &Val &CycleGAN &Ours\\
\midrule
C1: ``color/shoe" vs. ``edge/shoe"  &99 &0 &\bf{99}\\
C2: ``edge/handbag" vs. ``edge/shoe" &99 &\bf{99} &98\\
Both C1 and C2 &N/A &0 &\bf{98}\\
\bottomrule
\end{tabular}
}
\caption{The accuracy (\%) of classifying ``edge/shoe" images synthesized via ConceptGAN (ours) vs. CycleGAN~\cite{CycleGAN}. Joint classification accuracy is reported as the percentage of the images correctly classified in two tests at the same time. 
}
\label{tab:res1}
\end{table}


\begin{table}
\centering
\scalebox{0.92}{
\begin{tabular}{lccc}
\toprule
Classifier &Val&CycleGAN &Ours\\
\midrule
C1: ``with" vs. ``no" eyeglasses &98&93 &\bf{98}\\
C2: ``with" vs. ``no" bangs &93&61 &\bf{67}\\
Both C1 and C2 &N/A&56 &\bf{66}\\
\bottomrule
\end{tabular}
}
\caption{The accuracy (\%) of classifying face images synthesized via ConceptGAN (ours) vs. CycleGAN~\cite{CycleGAN}. 
}
\label{tab:res2}
\end{table}

\begin{table*}
\centering
\scalebox{0.95}{
\begin{tabular}{|c|c|c|c|c|c|c|c|c|c|}
\hline
Attributes &\multicolumn{3}{c|}{Smiling \& Eyeglasses} &\multicolumn{3}{c|}{Bangs \& Eyeglasses} & \multicolumn{3}{c|}{Smiling, Bangs, \& Eyeglasses} \\
\hline
Ranking Method &$l_{2}$ & RNP &SRID &$l_{2}$ & RNP &SRID &$l_{2}$&RNP &SRID\\
\cline{1-10}
Augmentation & No & Yes & Yes & No&Yes &Yes &No&Yes & Yes \\
\hline\hline
CaffeFace & 8.3& 10.7 &12.8 &7.9& 12.3& 16.9 &11.5 & 13.3 &16.6  \\ \hline
VGGFace & 38.6& 43.9 & 49.4 &49.8 & 59.4& 61.5 &44.4 & 54.8 &58.6\\ \hline
\end{tabular}
}
\caption{Rank-1 face verification results (in \%) for three different attribute sets: no augmentation (where we use $l_{2}$ distance to rank) vs. augmentation with ConceptGAN (where we use the multi-shot ranking algorithms, RNP and SRID to rank).}
\label{tab:res3}
\end{table*}

Key results of this experiment for multiple concept examples include the following. (a) \textbf{Classifying shoe and edge images:} In this experiment, we demonstrate results on the  ``handbag vs. shoe" and ``color vs. edge" concepts. We use images of ``color/handbag" ($\Sigma_{00}$), ``color/shoe" ($\Sigma_{10}$), and ``edge/handbag" ($\Sigma_{01}$) for learning the mappings of our proposed concept learning approach as well as individual mappings for CycleGAN. 
The results of this experiment are shown in Table~\ref{tab:res1}. The results demonstrate that the proposed method successfully composes two concepts in the subdomain $\Sigma_{11}$ as 98$\%$ of the synthesized images pass both classification tests, which greatly outperforms the results of direct composition of two separately trained CycleGAN units where no synthesized image survive both tests. (b) \textbf{Classifying face images with ``eyeglasses" and ``bangs":} In this experiment, we demonstrate results on the ``eyeglasses" and ``bangs" concepts. We use images of ``no eyeglasses, no bangs" ($\Sigma_{00}$), ``with eyeglasses, no bangs" ($\Sigma_{10}$), and ``no eyeglasses, with bangs" ($\Sigma_{01}$) to learn the mappings of ConceptGAN  and baseline CycleGAN. 
The results of this experiment are shown in Table~\ref{tab:res2}. The proposed method outperforms direct composition of CycleGAN units in terms of the synthesis quality in the subdomain $\Sigma_{11}$ by around $10\%$ improvement in joint classification accuracy.

\subsection{Face verification with augmented data}

Given a pair of face images, face verification is the problem of determining whether the pair represents the same person. Here, we demonstrate the applicability of ConceptGAN to this problem. Specifically, we begin with the one-shot version where every person in the probe and the gallery has exactly one image each. We then use the learned concept mappings to synthesize new, unseen face images, transforming the one-shot version to a multi-shot one. We demonstrate that by performing this conversion with our synthesized images, we improve the face verification performance. Here, we note that the focus of these evaluations is not to obtain state-of-the-art results but to demonstrate the applicability of ConceptGAN as a plug-in module that can be used in conjunction with any existing face verification algorithm to obtain improved performance. We use the CelebA~\cite{celebA} dataset for all experiments, where we generate 10 random splits of 100 people each not used in training ConceptGAN and report the average performance. 

\begin{figure}
\begin{center}
\includegraphics[scale=0.5]{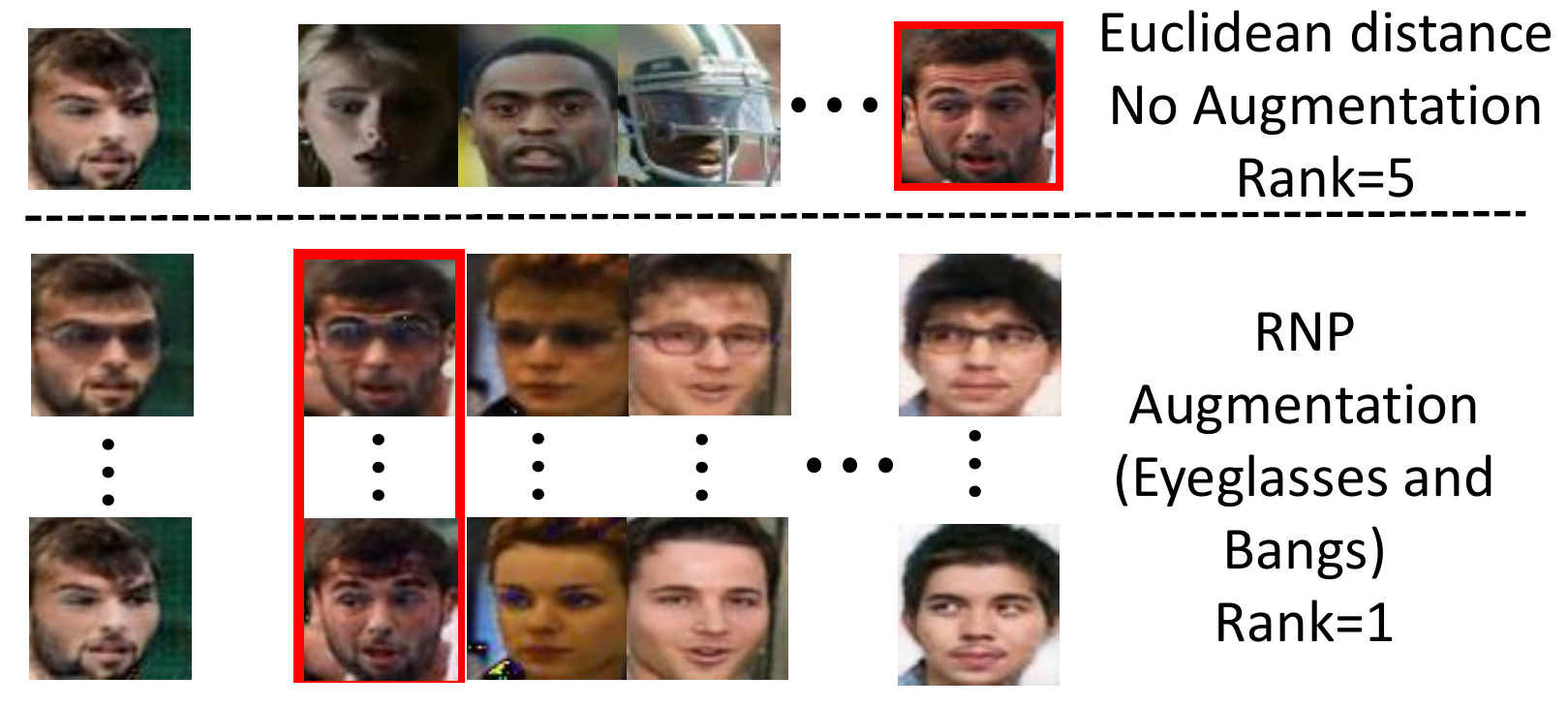}
\end{center}
   \caption{A qualitative illustration of improvement in face verification performance with augmented data using the ``eyeglasses" and ``bangs" attribute pair.}
\vspace{-1.3em}
\label{fig:rankImprovement}
\end{figure}

\begin{figure}
\begin{center}
   \includegraphics[scale=0.45]{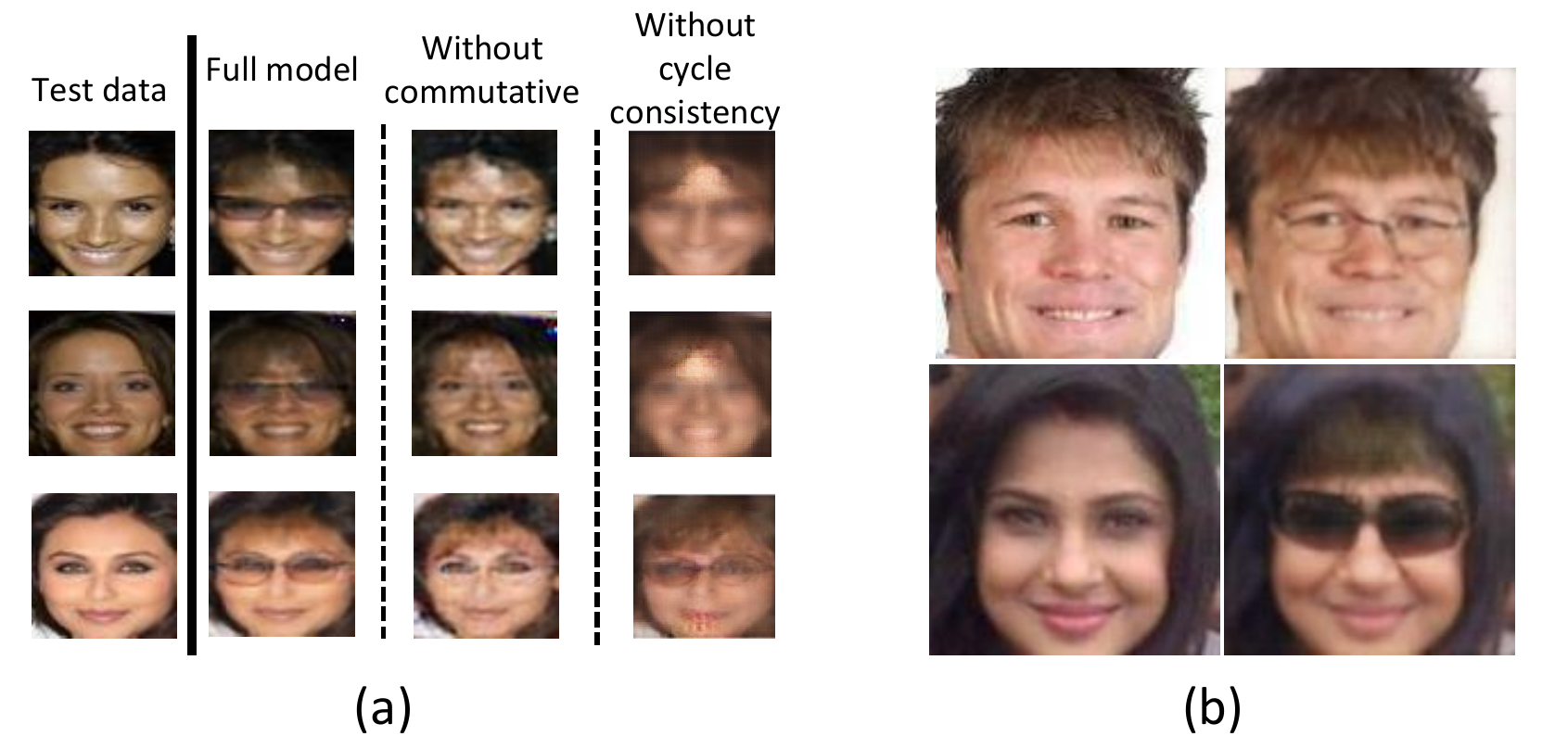}
\end{center}
\vspace{-2.3em}
\caption{(a) Sample qualitative results for ablation experiments. (b) Synthesis results for 128x128 images with ``Bangs" and "Eyeglasses": column1: $\Sigma_{00}$, column2: synthesized in $\Sigma_{11}$.}
\label{fig:ablationFig}
\end{figure}

\begin{table}
\centering
\begin{tabular}{lcc}
\toprule
Ranking method & $l_{2}$ & SRID \\
\toprule
Augmentation &No & Yes \\
\toprule
LFW &9.5 &13.1 \\
MS-Celeb1M  &11.7 &14.8\\
\bottomrule
\end{tabular}
\vspace{-1em}
\caption{Rank-1 face verification results (in \%): Transfer of concepts learned on CelebA to LFW and MS-Celeb1M.}
\label{tab:faceTransferResults}
\end{table}

\begin{figure}[!t]
\begin{center}
\includegraphics[scale=0.425]{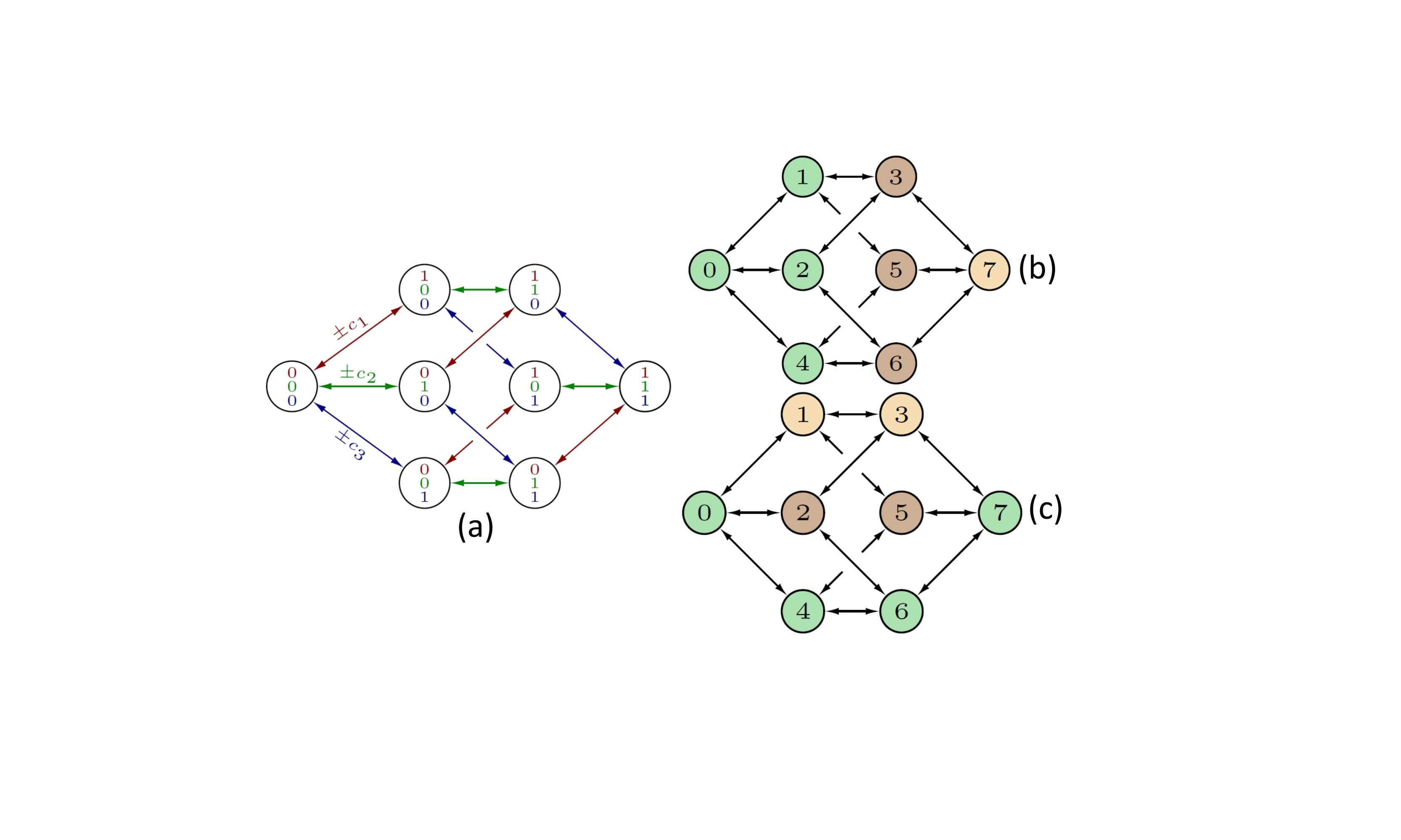}
\end{center}
   \caption{Generalizing ConceptGAN to $n$ concepts, illustrated with $n=3$. (a) concepts $c_1, c_2, c_3$, the $2^n=8$ states, and all possible shifts between the states. (b) $c_1, c_2, c_3$ defined by observing nodes 0,1,2,4, allowing primary inference of nodes 3,5,6, and secondary inference of node 7; (c) $c_1, c_2, c_3$ defined by observing nodes 0,4,6,7 with resulting primary (brown) and secondary (sand) inferred nodes.}
\vspace{-2em}
\label{fig:gen}
\end{figure}

We first perform one-shot experiments where we use two popular pre-trained face representation models, VGGFace \cite{vggface} and CaffeFace \cite{caffeface} to compute feature representations of the images and rank gallery candidates using the Euclidean distance. We next perform multi-shot experiments by augmenting both probe and gallery sets for each person using ConceptGAN, and rank gallery candidates with two multi-shot ranking algorithms, SRID \cite{srid,karanam2016systematic} and RNP \cite{rnp}. Results of all the experiments discussed above are summarized in Table~\ref{tab:res3}, where the augmented probe and gallery sets have 4 images each in the cases of two concepts and 8 images each in the case of 3 concepts. As can be noted from these results, converting the one-shot face verification problem to a multi-shot one by means of ConceptGAN has obvious benefits, with the multi-shot rank-1 face verification results consistently outperforming the corresponding one-shot results. We further qualitatively show the rank improvement in Figure~\ref{fig:rankImprovement}, where we see improved retrieval in the cases where face verification was performed with augmented data. 
Here we also provide quantitative evaluations for the transferability of concepts learned by CycleGAN. Specifically, in Table~\ref{tab:faceTransferResults}, we show rank-1 face verification results with CaffeFace and SRID on two independent test sets (LFW and MS-Celeb1M) using concepts learned by ConceptGAN on the CelebA dataset, where we see improved performance with data synthetized using the transferred concepts.  These results, complemented by the qualitative evaluations of the previous section, provide evidence for the transferability of the learned concepts to new datasets, demonstrating promise in learning the underlying latent space information.


\subsection{Ablation experiments}
In this section, we study the impact of the various components of the proposed loss function presented in Section~\ref{sec:overallLoss}. We first present qualitative results in Figure~\ref{fig:ablationFig}(a) for sample test images with and without the commutative ($\mathcal{L}_{comm}$) and distance-4 cycle consistency loss ($\mathcal{L}_{cyc4}$). In each case, we start with a test image (``No Bangs" and ``No Eyeglasses") and show the synthesized image in subdomain $\Sigma_{11}$ (with ``Bangs" and ``Eyeglasses"). One can see that with the ``full model", the visual quality of the generated images is better. We note that with SRID and CaffeFace, we obtain a rank-1 face verification performance of $12.1\%$ without $\mathcal{L}_{cyc4}$, $15.3\%$ without $\mathcal{L}_{comm}$, and $16.9\%$ with the full model. Furthermore, we report the following attribute classification results corresponding to those in Table~\ref{tab:res2}: combined C1 and C2 performance of $18\%$ without $\mathcal{L}_{cyc4}$, $60\%$ without $\mathcal{L}_{comm}$, and $66\%$ with the full model. These results provide empirical justification for each of the individual terms in our proposed loss function presented in Section~\ref{sec:overallLoss}. Finally, we also provide sample results for synthesizing images of resolution higher than the $64\times 64$ discussed previously- in Figure~\ref{fig:ablationFig}(b), we provide two examples of synthesizing $128\times 128$ images using our model. Additional results can be found in the supplementary material \footnote{Supplementary material can be found at \url{https://arxiv.org/abs/1711.06148}}. 

\section{Generalizing to multiple concepts}
In the previous sections, we discussed a possible way we could scale up to three concepts, and showed qualitative and quantitative results. Here, we provide a scheme to generalize our method to $n$ concepts under two assumptions: (a) concepts have distinct states, i.e.\ they are not continuous, and (b) activating one concept does not inhibit any other. We show that pairwise constraints over two concepts are sufficient for generating samplers from all concept combinations. Figure~\ref{fig:gen}(a) illustrates $n=3$ with concepts $\mathbb{C}=\{c_1, c_2, c_3\}$ as a graph where the edges apply a concept and the nodes are the $2^n$ concept combinations. Each node of the graph may be observed or not as illustrated in figure~\ref{fig:gen}(b) (green indicates an observed node). ``Observed" means that we have samples from the underlying distribution of a node. Applying our method then allows to infer node $3$, indicated in brown, with two concepts $\pm c_1$ and $\pm c_2$ involved. Indeed, the sub-graph of nodes $\{0,1,2,3\}$ is exactly our proposed two concept solution. Let's add data drawn from node $4$, observing the additional concept $\pm c_3$. The resulting graph shows that we can also infer nodes $5$ and $6$ by adding constraints corresponding the cycles $(0,2,6,4)$ and $(0,1,5,4)$. We now take the next step in generalization by considering node $7$. Assuming that we indeed can infer nodes ${3,5,6}$, we consider constraints that treat them as ``observed", such as over the cycles $(3,7,5,1)$, $(5,7,6,4)$, and $(6,7,3,2)$. This allows us to estimate samples for node $7$. To illustrate the generic nature, figure~\ref{fig:gen}(c) shows a situation with data at nodes $\{0,4,6,7\}$. We can firstly infer nodes $\{2,5\}$ and secondarily $\{1,3\}$. Generalizing to $n>3$, we propose to discover new layers of nodes in order of their distance from any observed node. Naturally, one cannot escape the combinatorial complexity of generating all the samplers. However, our generalization paves the way for iterative algorithms that yield approximate solutions efficiently based on a graphical representation of concepts and data.

\section{Conclusions}

We proposed ConceptGAN, a novel concept learning framework where we seek to capture underlying semantic shifts between data domains instead of mappings restricted to training distributions. The key idea is that via joint concept learning, transfer and composition, information over a joint latent space is recovered given incomplete training data. We showed that the proposed method can be applied as a smart data augmentation technique to generate realistic samples over different variations of concept attributes, including samples in a subdomain where the variation is completely unseen at the training stage. We demonstrated the compositionality of the captured concepts as well as the transferability of data augmentation in application on face verification problems. 

{\small
\bibliographystyle{ieee}
\bibliography{conceptLearning}
}

\end{document}